\documentclass[letterpaper]{article} 
\usepackage{aaai23}  
\usepackage{times}  
\usepackage{helvet}  
\usepackage{courier}  
\usepackage[hyphens]{url}  
\usepackage{graphicx} 
\usepackage{natbib}  
\usepackage{caption} 
\usepackage{booktabs}
\usepackage{multirow}
\usepackage{svg}
\usepackage{amsmath,amssymb}
\usepackage{here}
\usepackage{url}
\usepackage{comment}
\usepackage{multirow}
\usepackage{tabularx}

\usepackage{newfloat}
\usepackage{listings}
\DeclareCaptionStyle{ruled}{labelfont=normalfont,labelsep=colon,strut=off} 
\floatstyle{ruled}
\newfloat{listing}{tb}{lst}{}
\floatname{listing}{Listing}
\title{Penalizing Confident Predictions on Largely Perturbed Inputs Does Not Improve Out-of-Distribution Generalization in Question Answering}
\author{
    Kazutoshi Shinoda\textsuperscript{\rm 1,2},
    Saku Sugawara\textsuperscript{\rm 2},
    Akiko Aizawa\textsuperscript{\rm 1,2}
}
\affiliations{
    \textsuperscript{\rm 1}The University of Tokyo\\
    \textsuperscript{\rm 2}National Institute of Informatics\\
    shinoda@is.s.u-tokyo.ac.jp, \{saku, aizawa\}@nii.ac.jp
}

\begin{document}

\maketitle

\begin{abstract}
Question answering (QA) models are shown to be insensitive to large perturbations to inputs; that is, they make correct and confident predictions even when given largely perturbed inputs from which humans can not correctly derive answers.
In addition, QA models fail to generalize to other domains and adversarial test sets, while humans maintain high accuracy.
Based on these observations, we assume that QA models do not use intended features necessary for human reading but rely on spurious features, causing the lack of generalization ability.
Therefore, we attempt to answer the question: If the overconfident predictions of QA models for various types of perturbations are penalized, will the out-of-distribution (OOD) generalization be improved?
To prevent models from making confident predictions on perturbed inputs, we first follow existing studies and maximize the entropy of the output probability for perturbed inputs.
However, we find that QA models trained to be sensitive to a certain perturbation type are often insensitive to unseen types of perturbations.
Thus, we simultaneously maximize the entropy for the four perturbation types (i.e., word- and sentence-level shuffling and deletion) to further close the gap between models and humans.
Contrary to our expectations, although models become sensitive to the four types of perturbations, we find that the OOD generalization is not improved.
Moreover, the OOD generalization is sometimes degraded after entropy maximization.
Making unconfident predictions on largely perturbed inputs per se may be beneficial to gaining human trust.
However, our negative results suggest that researchers should pay attention to the side effect of entropy maximization.
\end{abstract}

\section{Introduction}
\label{sec:introduction}
Pretrained language models \cite{bert,liu2019roberta,lewis-etal-2020-bart,radford2019language,brown-etal-2020-language} have achieved human-level performance on natural language understanding (NLU) tasks, such as question answering (QA) \cite{rajpurkar-etal-2016-squad}, and natural language inference (NLI) \cite{williams-etal-2018-broad}.
Additionally, recent studies have shown that pretrained language models capture linguistic features based on an analysis with probing classifiers \cite{tenney2018what,hewitt-manning-2019-structural,hewitt-liang-2019-designing,belinkov-2022-probing}.

\begin{table}[t]
    \centering
    \small
    \begin{tabular}{p{0.9cm}p{6.5cm}}
    \toprule
    \multicolumn{2}{l}{\emph{Original input}}\\
    \midrule
    Context & The American Football Conference (AFC) champion \textbf{Denver Broncos} defeated the National Football Conference (NFC) champion Carolina Panthers 24–10 to earn their third Super Bowl title. \\
    \midrule
    Question & Which NFL team represented the AFC at Super Bowl 50? \\\midrule
    \multicolumn{2}{l}{\emph{Perturbed with function word deletion}}\\
    \midrule
    Context & American Football Conference AFC champion \textbf{Denver Broncos} defeated National Football Conference NFC champion Carolina Panthers 24 earn third Super Bowl title. \\\midrule
    Question & NFL team represented AFC Super Bowl 50? \\
    \midrule
    \multicolumn{2}{l}{\emph{Perturbed with word order shuffling}}\\
    \midrule
    Context & an Carolina the Super 10 American The National third their defeated NFC Conference champion Football to \textbf{Denver Broncos} 24 AFC ( Panthers ( champion. \\\midrule
    Question & at represented NFL team the AFC 50 Which Bowl Super? \\
    \bottomrule
    \end{tabular}
    \caption{Examples of largely perturbed inputs taken from SQuAD. In word order shuffling, we ensure that the answer spans indicated by \textbf{bold} remain as they are.}
    \label{tb:example}
\end{table}

\begin{table*}[tbp]
    \centering
    \begin{tabular}{lll}
    \toprule
    Perturbation $\sigma$ & Description & Intended feature removed by perturbation \\
    \midrule
    ${\rm Del_{func}}$ &  Delete all the function words & Function words\\
    ${\rm Del_{que}}$ & Delete the question & Question words\\
    ${\rm Shuf_{word}}$ & Shuffle the word order in each sentence & Syntactic information \\
    ${\rm Shuf_{sent}}$ & Shuffle the sentence order in a context & Discourse relations\\
    \bottomrule
    \end{tabular}
    \caption{Four types of perturbations $\sigma$ studied in this work. Different perturbations remove different types of intended features necessary for human reading from the inputs.}
    \label{tb:perturbation}
\end{table*}

However, whether language models fine-tuned on QA tasks have human-like language understanding abilities remains debatable.
QA models often maintain high accuracy and confidence scores even when the inputs are transformed by large perturbations (e.g., word deletion \cite{feng-etal-2018-pathologies,sugawara-etal-2018-makes}, word order shuffling, sentence deletion, and sentence order shuffling \cite{sugawara-etal-2020-assessing}).
See Table \ref{tb:example} for the examples of largely perturbed inputs in extractive QA.
Similar phenomena have been observed in NLI \cite{sinha-etal-2021-unnatural}, and other NLU tasks \cite{gupta_bert_2021}.
Meanwhile, our human evaluation (\S\ref{sec:human-evaluation}) indicates that humans cannot correctly derive answers from such invalid inputs.
We argue that such phenomena imply that models do not adequately use semantic and syntactic features removed by perturbations to make predictions, while these features are indispensable for humans to understand language.

In addition, QA models are shown to lack out-of-distribution generalization.
QA models trained on a certain dataset fail to generalize to datasets from other domains \cite{yogatama2019learning,talmor-berant-2019-multiqa,sen-saffari-2020-models}.
They also lack robustness to adversarial attacks that append fake sentences to contexts \cite{jia-liang-2017-adversarial}.

Given the two characteristics of QA models (i.e., the insensitivity to large perturbations and the lack of generalization ability), we aim to answer the following question: \emph{If the overconfident predictions of QA models for various types of perturbations are penalized, will the out-of-distribution (OOD) generalization be improved?}
Previous studies have shown that maximizing the entropy \cite{shannon1948mathematical} of the output probability for perturbed inputs can successfully reduce model confidence for such perturbed inputs \cite{feng-etal-2018-pathologies,sinha-etal-2021-unnatural,gupta_bert_2021}.
We adopt this method to make QA models sensitive to the perturbations listed in Table \ref{tb:perturbation}.

However, we observe that entropy maximization for a certain perturbation type can transfer to the seen perturbation type but often fails to transfer to unseen perturbation types.
For example, after maximizing the entropy for question deletion, models are not sensitive to function word deletion.
To mitigate this lack of transferability, we propose to simultaneously maximize the entropy for the predefined perturbation types.
We show that this approach is effective to make models recognize all the predefined perturbations while maintaining in-domain accuracy.

Contrary to our expectations, even though models become sensitive to the four types of perturbations, we find that the generalization to other domains or adversarial robustness is not improved.
As discussed in \citet{hase2021search}, intentionally perturbed inputs become unnatural and are unlikely to appear in a dataset.
Therefore, making models sensitive to largely perturbed inputs may a have negative impact on out-of-distribution generalization.
While becoming sensitive to unnatural inputs with entropy maximization like humans can gain trust from humans, our results suggest that researchers should pay attention to the side effect of entropy maximization.

Our main contributions are as follows:
\begin{itemize}
    \item We find that entropy maximization can mitigate the insensitivity to seen perturbation types, but fail to transfer to unseen perturbation types in QA.
    \item We show that simply maximizing the entropy for the four perturbation types, including word- and sentence-level ones, can mitigate this issue.
    \item We show that even though QA models become sensitive to the four types of perturbations, the generalization to other domains or adversarial robustness is not improved but rather sometimes degraded.
\end{itemize}

\section{Method}
\label{method}
\subsection{Perturbation Types}
We list the examined perturbations in Table \ref{tb:perturbation}.
We adopt two word-level perturbations, function word deletion (${\rm Del_{func}}$) and word order shuffling (${\rm Shuf_{word}}$), and two sentence-level perturbations, question deletion (${\rm Del_{que}}$) and sentence order shuffling (${\rm Shuf_{sent}}$) to comprehensively assess the sensitivity of QA models to the intended features necessary for humans to understand language, which cover the surface structure and the textbase of the construction--integration model comprehensively.
We adopted these perturbations because a QA model is relatively insensitive to them compared to other types of perturbations (e.g., contend word deletion and vocabulary anonymization) as found in \citet{sugawara-etal-2020-assessing}.
We expect that entropy maximization with these perturbations make models learn to recognize the intended features as shown in Table \ref{tb:perturbation}.
The detailed motivation of the expectation is described in \S\ref{sec:interpretation}.

\subsection{Entropy Maximization}
To penalize the confident predictions of models on perturbed inputs, we adopt entropy maximization used by \citet{feng-etal-2018-pathologies,gupta_bert_2021}.
Namely, we minimize the cross-entropy loss while maximizing the entropy of the output probabilities given perturbed inputs.

When the dataset $\mathcal{D}$ consists of pairs of input $x$ and output $y$, and the model parameters are $\theta$, the cross-entropy loss is given by
\begin{equation}
\label{eq:cross}
\mathcal{L}_{ce} = - \frac{1}{|\mathcal{D}|} \sum_{(x, y) \in \mathcal{D}} \log p_\theta(y|x).
\end{equation}
\noindent When a perturbed input $x_{\sigma}$ is obtained from $x$ by applying a perturbation $\sigma$, the entropy of the model output given perturbed input is given by\footnote{Uppercase letters (e.g., $X$) represent random variables and lowercase letters (e.g., $x$) represent actual values.}:
\begin{align}
\label{eq:entropy}
H(Y|X_{\sigma}) = \frac{1}{|\mathcal{D}|} \sum_{(x,y) \in \mathcal{D}} - p_\theta(y|x_{\sigma}) \log p_\theta(y|x_{\sigma}).
\end{align}
\noindent The loss function to be minimized is computed as follows:
\begin{equation}
\label{eq:loss}
\mathcal{L} = \mathcal{L}_{ce} - \lambda_{\sigma} H(Y|X_{\sigma}).
\end{equation}
\noindent where the entropy term is scaled by the factor $\lambda_\sigma$ ($>0$).

\subsection{Conditional Independence Assumption for Extractive QA}
\label{app:ent-qa}
In extractive QA tasks, such as SQuAD \cite{rajpurkar-etal-2016-squad}, the models need to specify the start and end positions of the predicted answer span in the context for the given question.
When computing the conditional probability $p_\theta(y|x)$ in Equations \ref{eq:cross} and \ref{eq:entropy}, as most existing studies implicitly did during training \cite{seo2017bidirectional,bert}, we assume that the start and end positions of answers, i.e., $Y_{start}$ and $Y_{end}$, are conditionally independent given the context and question for brevity.
Namely, we assume that $p(Y|X)=p(Y_{start}|X)p(Y_{end}|X)$.
Based on this assumption, the entropy term in Equations \ref{eq:loss} and \ref{eq:all-loss} can be computed as follows:
\begin{equation}
\label{eq:relaxation}
H(Y|X_{\sigma}) = H(Y_{start}|X_{\sigma}) + H(Y_{end}|X_{\sigma}).
\end{equation}
\noindent We adopt this relaxation because it is costly to raise all the possible answer spans meeting the condition that the start position is lower than or equal to the end position.
Our experiments show that this does not degrade the in-distribution accuracy.\footnote{The entropy of output probabilities can be also defined in multiple-choice and abstractive QA. Extending our work to other QA formats is future work.}

\subsection{Recognizing Multiple Types of Perturbations}
Our experiments in \S\ref{sec:unseen-perturbation} show that maximizing entropy for a certain perturbation type does not transfer to unseen perturbation types.
We need to mitigate this problem because our aim is to investigate whether making models sensitive to the four types of perturbations in Table \ref{tb:perturbation} improves out-of-distribution generalization.

To mitigate the lack of transferabiliity, we propose to maximize the entropy term for the four type of perturbations to make models recognize those features as follows:
\begin{equation}
\label{eq:all-loss}
\mathcal{L} = \mathcal{L}_{ce} - \sum_{\sigma} \lambda_{\sigma} H(Y|X_{\sigma}).
\end{equation}

\begin{table*}[t]
    \centering
    \begin{tabular}{c|c|ccccc}
    \toprule
Model & 
\begin{tabular}{c}
     Perturbation \\
     train↓ / test→
\end{tabular}
 &  None  & ${\rm Del_{func}}$ & ${\rm Del_{que}}$ & ${\rm Shuf_{word}}$ & ${\rm Shuf_{sent}}$\\
\midrule
\multirow{6}{*}{\rotatebox{90}{BERT-base}} & None & 1.38{\scriptsize$\pm{0.00}$} & 3.43{\scriptsize$\pm{0.16}$} & 7.03{\scriptsize$\pm{1.10}$} & 3.53{\scriptsize$\pm{0.16}$} & 1.43{\scriptsize$\pm{0.00}$}\\
 & ${\rm Del_{func}}$ & 1.37{\scriptsize$\pm{0.00}$} & \textbf{11.9}{\scriptsize$\pm{0.00}$} & 7.1{\scriptsize$\pm{0.20}$} & 10.48{\scriptsize$\pm{0.36}$} & 1.41{\scriptsize$\pm{0.01}$}\\
 & ${\rm Del_{que}}$ & 1.37{\scriptsize$\pm{0.01}$} & 3.69{\scriptsize$\pm{0.28}$} & \textbf{11.9}{\scriptsize$\pm{0.00}$} & 4.31{\scriptsize$\pm{0.21}$} & 1.41{\scriptsize$\pm{0.01}$}\\
 & ${\rm Shuf_{word}}$ & 1.37{\scriptsize$\pm{0.00}$} & 11.87{\scriptsize$\pm{0.01}$} & 7.43{\scriptsize$\pm{0.27}$} & \textbf{11.9}{\scriptsize$\pm{0.00}$} & 1.41{\scriptsize$\pm{0.00}$}\\
 & ${\rm Shuf_{sent}}$ & 1.43{\scriptsize$\pm{0.01}$} & 3.65{\scriptsize$\pm{0.04}$} & 7.94{\scriptsize$\pm{0.35}$} & 3.82{\scriptsize$\pm{0.23}$} & \textbf{1.48}{\scriptsize$\pm{0.01}$}\\
 & ALL & 1.41{\scriptsize$\pm{0.01}$} & \textbf{11.9}{\scriptsize$\pm{0.00}$} & \textbf{11.9}{\scriptsize$\pm{0.00}$} & \textbf{11.9}{\scriptsize$\pm{0.00}$} & 1.47{\scriptsize$\pm{0.01}$}\\
\midrule
\multirow{6}{*}{\rotatebox{90}{RoBERTa-base}} & None & 1.13{\scriptsize$\pm{0.05}$} & 3.17{\scriptsize$\pm{0.85}$} & 8.29{\scriptsize$\pm{0.39}$} & 2.76{\scriptsize$\pm{0.48}$} & 1.45{\scriptsize$\pm{0.02}$}\\
 & ${\rm Del_{func}}$ & 1.10{\scriptsize$\pm{0.02}$} & \textbf{11.9}{\scriptsize$\pm{0.0}$} & 9.05{\scriptsize$\pm{0.19}$} & 11.89{\scriptsize$\pm{0.00}$} & 1.44{\scriptsize$\pm{0.04}$}\\
 & ${\rm Del_{que}}$ & 1.12{\scriptsize$\pm{0.03}$} & 2.94{\scriptsize$\pm{0.47}$} & \textbf{11.9}{\scriptsize$\pm{0.00}$} & 2.59{\scriptsize$\pm{0.32}$} & 1.42{\scriptsize$\pm{0.01}$}\\
 & ${\rm Shuf_{word}}$ & 1.13{\scriptsize$\pm{0.03}$} & 8.95{\scriptsize$\pm{2.74}$} & 8.51{\scriptsize$\pm{0.31}$} & \textbf{11.9}{\scriptsize$\pm{0.00}$} & 2.38{\scriptsize$\pm{0.28}$}\\
 & ${\rm Shuf_{sent}}$ & 1.09{\scriptsize$\pm{0.01}$} & 2.27{\scriptsize$\pm{0.59}$} & 9.08{\scriptsize$\pm{0.23}$} & \textbf{11.9}{\scriptsize$\pm{0.00}$} & \textbf{11.9}{\scriptsize$\pm{0.00}$}\\
 & ALL & 1.14{\scriptsize$\pm{0.04}$} & \textbf{11.9}{\scriptsize$\pm{0.00}$} & \textbf{11.9}{\scriptsize$\pm{0.00}$} & \textbf{11.9}{\scriptsize$\pm{0.00}$} & \textbf{11.9}{\scriptsize$\pm{0.00}$}\\
\bottomrule
\end{tabular}
    \caption{Entropy of the model predictions on the original and perturbed SQuAD 1.1 development set. The more confident predictions models make, the lower entropy is.}
    \label{tb:entropy}
\end{table*}

\begin{table*}[h]
    \centering
    \begin{tabular}{c|c|ccccc}
    \toprule
    Model &
    \begin{tabular}{c}
        Perturbation \\
        train↓ / test→
    \end{tabular}
& None & ${\rm Del_{func}}$ & ${\rm Del_{que}}$ & ${\rm Shuf_{word}}$ & ${\rm Shuf_{sent}}$ \\
\midrule
\multirow{6}{*}{\rotatebox{90}{BERT-base}} & None & 88.0{\scriptsize $\pm{0.03}$} & 54.2{\scriptsize $\pm{0.06}$} & 10.2{\scriptsize $\pm{0.41}$} & 26.5{\scriptsize $\pm{0.14}$} & 83.9{\scriptsize $\pm{0.06}$}\\
 & ${\rm Del_{func}}$ & 88.1{\scriptsize $\pm{0.02}$} & \textbf{22.2}{\scriptsize $\pm{3.83}$} & 10.2{\scriptsize $\pm{0.28}$} & 24.2{\scriptsize $\pm{0.73}$} & 83.8{\scriptsize $\pm{0.26}$}\\
 & ${\rm Del_{que}}$ & 88.1{\scriptsize $\pm{0.12}$} & 53.9{\scriptsize $\pm{0.91}$} & \textbf{5.9}{\scriptsize $\pm{0.74}$} & 26.4{\scriptsize $\pm{0.36}$} & 84.1{\scriptsize $\pm{0.14}$}\\
 & ${\rm Shuf_{word}}$ & 88.1{\scriptsize $\pm{0.07}$} & 36.4{\scriptsize $\pm{0.32}$} & 10.0{\scriptsize $\pm{0.37}$} & \textbf{16.2}{\scriptsize $\pm{1.53}$} & 83.8{\scriptsize $\pm{0.25}$}\\
 & ${\rm Shuf_{sent}}$ & 88.0{\scriptsize $\pm{0.09}$} & 54.3{\scriptsize $\pm{0.79}$} & 9.9{\scriptsize $\pm{0.53}$} & 26.8{\scriptsize $\pm{0.29}$} & 83.9{\scriptsize $\pm{0.18}$}\\
 & ALL & 88.0{\scriptsize $\pm{0.10}$} & 31.1{\scriptsize $\pm{2.61}$} & 7.9{\scriptsize $\pm{1.81}$} & 19.1{\scriptsize $\pm{0.41}$} & 83.9{\scriptsize $\pm{0.14}$}\\
\midrule
\multirow{6}{*}{\rotatebox{90}{RoBERTa-base}} & None & 91.2{\scriptsize $\pm{0.04}$} & 61.0{\scriptsize $\pm{0.72}$} & 11.3{\scriptsize $\pm{0.33}$} & 29.3{\scriptsize $\pm{0.06}$} & 87.3{\scriptsize $\pm{0.21}$}\\
 & ${\rm Del_{func}}$ & 91.4{\scriptsize $\pm{0.01}$} & \textbf{14.5}{\scriptsize $\pm{2.21}$} & 11.0{\scriptsize $\pm{0.21}$} & 19.2{\scriptsize $\pm{0.88}$} & 87.4{\scriptsize $\pm{0.12}$}\\
 & ${\rm Del_{que}}$ & 91.2{\scriptsize $\pm{0.13}$} & 60.9{\scriptsize $\pm{0.53}$} & \textbf{7.0}{\scriptsize $\pm{2.44}$} & 28.9{\scriptsize $\pm{0.41}$} & 87.5{\scriptsize $\pm{0.12}$}\\
 & ${\rm Shuf_{word}}$ & 91.2{\scriptsize $\pm{0.17}$} & 47.8{\scriptsize $\pm{4.34}$} & 11.2{\scriptsize $\pm{0.12}$} & 12.1{\scriptsize $\pm{2.05}$} & 86.8{\scriptsize $\pm{0.30}$}\\
 & ${\rm Shuf_{sent}}$ & 91.3{\scriptsize $\pm{0.05}$} & 59.9{\scriptsize $\pm{0.50}$} & 10.2{\scriptsize $\pm{0.70}$} & 10.0{\scriptsize $\pm{1.87}$} & \textbf{17.0}{\scriptsize $\pm{5.06}$}\\
 & ALL & 91.3{\scriptsize $\pm{0.08}$} & 19.6{\scriptsize $\pm{3.74}$} & 8.9{\scriptsize $\pm{2.46}$} & \textbf{9.7}{\scriptsize $\pm{1.56}$} & 34.8{\scriptsize $\pm{7.65}$}\\
\midrule
\midrule
\multicolumn{2}{c|}{Human Score} & 91.2$^\dagger$ & 28.1 & 0.1 & 10.8 & 53.2 \\
\bottomrule
    \end{tabular}
    \caption{F1 scores on the original and perturbed SQuAD dev set. See Table \ref{tb:perturbation} for details of perturbation types. $^\dagger$Copied from the SQuAD 1.1 Leaderboard.}
    \label{tb:f1}
\end{table*}

\subsection{Interpretation from the Perspective of Causality}
\label{sec:interpretation}
When the maximum predicted probability (confidence score) of model $\theta$ for original input $x$ is $p_{\theta}(\hat{y}|x) = \textrm{max} ~ p_{\theta}(y|x)$, the difference in probabilities that the model assigns to $\hat{y}$,
\begin{equation}
\label{eq:diff}
d_{\theta}(x, \hat{y}, \sigma) = p(\hat{y}|x) - p(\hat{y}|x_{\sigma}),
\end{equation}
can be regarded as quantifying how much feature $s$ is used by the model for making prediction $\hat{y}$.
Quantities of similar definitions have been used as feature importance \cite{li2016understanding,deyoung-etal-2020-eraser,hase2021search} to increase the interpretability of the model, or the degree to which a cause affects an outcome in the context of causality \cite{pearl2000models}.

By minimizing the cross entropy while maximizing the entropy in Equation \ref{eq:loss}, $p(\hat{y}|x)$ is increased while $p(\hat{y}|x_{\sigma})$ is decreased.
Thus, minimizing $\mathcal{L}$ in Equation \ref{eq:loss} is expected to indirectly increase $d_{\theta}(x, \hat{y}, \sigma)$ in Equation \ref{eq:diff}.
When $d_{\theta}(x, \hat{y}, \sigma)$ is larger than non-zero values, features in $x$ removed by perturbation $\sigma$ have causal effects on prediction $\hat{y}$ made by QA model $\theta$.
Based on this interpretation, we assume that training with entropy maximization causes QA models to use intended features as listed in Table \ref{tb:perturbation}, and have positive impact on out-of-distribution generalization.

However, our experiments show that the results are opposite to the assumption.
We will discuss why the out-of-generalization is not improved with the approach in \S\ref{sec:ood}.

\section{Experiment}
\label{sec:experiment}
In this study, we considered a QA task because the inputs of QA datasets consist of questions and contexts, and the contexts often consist of multiple sentences.
This enables examination of broader types of perturbations, such as sentence order shuffling, that are absent in other NLU tasks such as NLI and paraphrase identification \cite{dolan-brockett-2005-automatically}, where the inputs are only two sentences.

\subsection{Experimental Setups}
\paragraph{Model} We used BERT-base \cite{bert} and RoBERTa-base \cite{liu2019roberta} for QA models because they are often adopted in QA.

\paragraph{Dataset} We used SQuAD 1.1 \cite{rajpurkar-etal-2016-squad} for training and evaluation.
To evaluate the generalization to other domains, we employed the dev set of NewsQA \cite{trischler-etal-2017-newsqa}, TriviaQA \cite{joshi-etal-2017-triviaqa}, SearchQA \cite{dunn2017searchqa}, HotpotQA \cite{yang-etal-2018-hotpotqa}, and NaturalQuestions \cite{kwiatkowski-etal-2019-natural} from MRQA 2019 shared task \cite{fisch-etal-2019-mrqa}.
To evaluate adversarial robustness, we used AddSent and AddOneSent from Adversarial SQuAD \cite{jia-liang-2017-adversarial}.

\paragraph{Training} We used the Adam \cite{kingma2014adam} optimizer with epsilon 1e-8.
The models were trained for two epochs with the learning rate being linearly decreased from 3e-5 to zero.
The batch size was set to 32.
For other hyperparameters, we generally used the default hyperparameters in the example code provided by Huggingface.
We tuned the scaling factor $\lambda_{\sigma}$ in Equation \ref{eq:loss} in $\{0.01, 0.1, 1.0, 5.0\}$ for each perturbation $\sigma$ on the SQuAD dev set based on the F1 scores.
The means and standard deviations of the F1 scores over three random seeds are reported.

\subsection{Human Evaluation}
\label{sec:human-evaluation}
To see whether humans can derive correct answers from inputs with the examined perturbations, we conducted a human evaluation.
We asked human annotators to answer a question by extracting an answer span from a given context.
The annotators are allowed to submit empty answers when they cannot find plausible answers.
The input is transformed by one of the four perturbation types.
We randomly chose 200 examples for each perturbation from the SQuAD dev set.
Three annotators on Amazon Mechanical Turk were assigned to each example.

\begin{table*}[tbp]
\centering
\begin{tabular}{c|c|ccccc}
\toprule
 Model & Perturbation & SearchQA & HotpotQA & NQ & NewsQA & TriviaQA \\
 \midrule
  \multirow{6}{*}{\rotatebox{90}{BERT-base}} & None & 27.3{\scriptsize $\pm{ 0.60 }$} & 60.6{\scriptsize $\pm{ 0.44 }$} & 59.1{\scriptsize $\pm{ 0.50 }$} & 55.8{\scriptsize $\pm{ 0.26 }$} & 58.5{\scriptsize $\pm{ 0.27 }$} \\
  & ${\rm Del_{func}}$ & 27.2{\scriptsize $\pm{ 0.98 }$} & 60.0{\scriptsize $\pm{ 0.37 }$} & 56.2{\scriptsize $\pm{ 0.39 }$} & 55.9{\scriptsize $\pm{ 0.37 }$} & 58.6{\scriptsize $\pm{ 0.19 }$} \\
  & ${\rm Del_{que}}$ & 27.4{\scriptsize $\pm{ 0.71 }$} & 60.0{\scriptsize $\pm{ 0.21 }$} & 58.7{\scriptsize $\pm{ 0.27 }$} & 55.5{\scriptsize $\pm{ 0.43 }$} & 58.6{\scriptsize $\pm{ 0.46 }$} \\
  & ${\rm Shuf_{word}}$ & 27.8{\scriptsize $\pm{ 0.29 }$} & 60.1{\scriptsize $\pm{ 0.02 }$} & 56.7{\scriptsize $\pm{ 0.42 }$} & 55.9{\scriptsize $\pm{ 0.52 }$} & 58.7{\scriptsize $\pm{ 0.16 }$} \\
  & ${\rm Shuf_{sent}}$ & 27.6{\scriptsize $\pm{ 1.83 }$} & 60.2{\scriptsize $\pm{ 0.20 }$} & 58.8{\scriptsize $\pm{ 0.45 }$} & 56.0{\scriptsize $\pm{ 0.15 }$} & 58.6{\scriptsize $\pm{ 0.08 }$} \\
  & ALL & 28.0{\scriptsize $\pm{ 1.08 }$} & 60.7{\scriptsize $\pm{ 0.45 }$} & 56.9{\scriptsize $\pm{ 0.81 }$} & 55.3{\scriptsize $\pm{ 0.44 }$} & 57.9{\scriptsize $\pm{ 0.49 }$} \\
  \midrule
  \multirow{6}{*}{\rotatebox{90}{RoBERTa-base}} & None & 30.7{\scriptsize $\pm{ 1.90 }$} & 66.5{\scriptsize $\pm{ 0.73 }$} & 61.8{\scriptsize $\pm{ 0.39 }$} & 64.3{\scriptsize $\pm{ 0.11 }$} & 62.7{\scriptsize $\pm{ 0.30 }$} \\
  & ${\rm Del_{func}}$ & 26.8{\scriptsize $\pm{ 2.49 }$} & 66.6{\scriptsize $\pm{ 0.25 }$} & 61.4{\scriptsize $\pm{ 0.22 }$} & 64.5{\scriptsize $\pm{ 0.21 }$} & 62.1{\scriptsize $\pm{ 0.58 }$} \\
  & ${\rm Del_{que}}$ & 31.5{\scriptsize $\pm{ 0.99 }$} & 66.2{\scriptsize $\pm{ 0.31 }$} & 62.0{\scriptsize $\pm{ 0.45 }$} & 64.7{\scriptsize $\pm{ 0.34 }$} & 62.8{\scriptsize $\pm{ 0.36 }$} \\
  & ${\rm Shuf_{word}}$ & 23.6{\scriptsize $\pm{ 1.65 }$} & 66.7{\scriptsize $\pm{ 0.48 }$} & 57.0{\scriptsize $\pm{ 2.39 }$} & 64.6{\scriptsize $\pm{ 0.07 }$} & 62.2{\scriptsize $\pm{ 0.37 }$} \\
  & ${\rm Shuf_{sent}}$ & 28.6{\scriptsize $\pm{ 1.03 }$} & 66.2{\scriptsize $\pm{ 0.42 }$} & 17.5{\scriptsize $\pm{ 3.90 }$} & 64.7{\scriptsize $\pm{ 0.27 }$} & 61.4{\scriptsize $\pm{ 0.43 }$} \\
  & ALL & 14.4{\scriptsize $\pm{ 3.51 }$} & 66.5{\scriptsize $\pm{ 0.81 }$} & 25.3{\scriptsize $\pm{ 4.56 }$} & 63.6{\scriptsize $\pm{ 0.24 }$} & 60.6{\scriptsize $\pm{ 0.38 }$} \\
\bottomrule
    \end{tabular}
    \caption{F1 scores on test sets in other domains. The means$\pm$standard deviations over three random seeds are reported.}
    \label{tb:ood-detail}
\end{table*}

\begin{table}[t]
    \centering
    \small
    \begin{tabular}{c|c|cc}
    \toprule
    Model & Perturbation & AddSent & AddOneSent \\
    \midrule
    \multirow{6}{*}{\rotatebox{90}{BERT-base}} & None & 50.8{\scriptsize $\pm{ 0.40 }$} & 62.1{\scriptsize $\pm{ 0.89 }$} \\
    & ${\rm Del_{func}}$ & 49.6{\scriptsize $\pm{ 0.54 }$} & 61.4{\scriptsize $\pm{ 1.26 }$} \\
    & ${\rm Del_{que}}$ & 50.7{\scriptsize $\pm{ 0.88 }$} & 62.2{\scriptsize $\pm{ 0.39 }$} \\
    & ${\rm Shuf_{word}}$ & 49.9{\scriptsize $\pm{ 0.63 }$} & 61.8{\scriptsize $\pm{ 1.14 }$} \\
    & ${\rm Shuf_{sent}}$ & 49.9{\scriptsize $\pm{ 0.87 }$} & 61.6{\scriptsize $\pm{ 1.08 }$} \\
    & ALL & 50.7{\scriptsize $\pm{ 0.71 }$} & 62.2{\scriptsize $\pm{ 0.74 }$} \\
    \midrule
    \multirow{6}{*}{\rotatebox{90}{RoBERTa-base}} & None & 62.6{\scriptsize $\pm{ 0.90 }$} & 72.0{\scriptsize $\pm{ 0.95 }$}\\
    & ${\rm Del_{func}}$ & 62.4{\scriptsize $\pm{ 1.00 }$} & 71.6{\scriptsize $\pm{ 1.33 }$}\\
    & ${\rm Del_{que}}$ & 61.9{\scriptsize $\pm{ 0.94 }$} & 71.6{\scriptsize $\pm{ 0.84 }$} \\
    & ${\rm Shuf_{word}}$ & 61.5{\scriptsize $\pm{ 0.37 }$} & 70.8{\scriptsize $\pm{ 0.65 }$} \\
    & ${\rm Shuf_{sent}}$ & 62.2{\scriptsize $\pm{ 1.61 }$} & 71.6{\scriptsize $\pm{ 1.10 }$} \\
    & ALL & 61.9{\scriptsize $\pm{ 1.11 }$} & 71.4{\scriptsize $\pm{ 0.41 }$} \\
    \bottomrule
    \end{tabular}
    \caption{F1 scores on adversarial test sets. The means$\pm$standard deviations over three random seeds are reported.}
    \label{tb:adversarial-robustness}
\end{table}

\subsection{Cross-Perturbation Evaluation}
\label{sec:unseen-perturbation}
Previous studies have shown that maximizing entropy for input with certain perturbations, such as word deletion \cite{feng-etal-2018-pathologies} and word order shuffling \cite{sinha-etal-2021-unnatural}, can make models sensitive (i.e., less confident) to the \textit{same} type of perturbations at test time.
Intuitively, the word orders and words themselves should convey different types of information.
Then, we can ask the question: Can maximizing the entropy for word order shuffling transfer to that for word deletion?

To answer this question, we investigated the transferability of entropy maximization across four types of perturbations. 
That is, we trained the QA models with entropy maximization for one of the four perturbations and evaluated the models with seen and unseen perturbations.
To evaluate the sensitivity of the predictions to the perturbations, we used the entropy of the predicted answers and the F1 score.\footnote{In this experiment, because the maximum length of the context is set to 384, the maximum of the entropy is 11.9 ($= - 1 / 384 * \log(1/384) * 384 * 2$), and the minimum is 0.0 based on Equations \ref{eq:entropy} and \ref{eq:relaxation}.}
The entropies and F1 scores for the in-distribution dev set are shown in Tables \ref{tb:entropy} and \ref{tb:f1}, respectively.

\paragraph{QA Models Are More Insensitive to Large Perturbations Than Humans}
First, without entropy maximization, models can somehow correctly answer decent portions of perturbed inputs compared with humans.
Notably, QA models trained without entropy maximization were most robust to ${\rm Shuf_{sent}}$ among the four perturbation types.
This result is consistent with the findings of \citet{sugawara-etal-2020-assessing}.
While the F1 scores decreased substantially and the entropy is relatively high when the inputs in the dev set are perturbed with ${\rm Del_{que}}$, the scores of the models are still higher than the human score.

These relatively high F1 scores imply that models can not recognize the intended features removed by the perturbations as humans do.
Moreover, the confident predictions of the models on invalid data may harm the reliability of model predictions in real-world applications.

\paragraph{Entropy Maximization Can Penalize Confident and Correct Predictions for Seen Perturbations} 
Entropy maximization make models sensitive to seen perturbation types, as shown by the diagonals of Tables \ref{tb:entropy} and \ref{tb:f1}, except for BERT-base with ${\rm Shuf_{sent}}$.
The F1 scores decreased and the entropies increased along the diagonal cells.
Moreover, the entropies for the seen perturbations almost reached the maximum value of 11.9 without hurting the F1 scores on the original dev set (None).

\paragraph{Entropy Maximization Fails to Transfer to Unseen Perturbations}
However, maximizing entropy for a certain perturbation type often cannot transfer to unseen perturbation types.
For example, maximizing entropy for ${\rm Del_{que}}$ have little impact on the sensitiveness of models to ${\rm Del_{func}}$.
To mitigate the lack of cross-perturbation transferability, we simply maximize the entropy terms for all the perturbation types as in Equation \ref{eq:all-loss}, which is denoted as ALL.
We show that this approach can successfully make models less confident on all the four perturbations except for BERT-base with ${\rm Shuf_{sent}}$.
On the other hand, we observed that there are some perturbation types where entropy maximization can transfer to some extent (e.g., from ${\rm Del_{func}}$ to ${\rm Shuf_{word}}$ and from ${\rm Shuf_{sent}}$ to ${\rm Shuf_{word}}$).

\paragraph{Influence of the Scaling Factor}
Among the perturbation types, BERT-base failed to become less confident on ${\rm Shuf_{sent}}$.
To determine the cause, we examined the influence of scaling factors.
First, the chosen scaling factor was 0.01, which was insufficient to increase the entropy.
Second, we found that the F1 scores of BERT-base on the clean dev set and the dev set perturbed with ${\rm Shuf_{sent}}$ are strongly correlated.
This implies that BERT-base cannot distinguish the original sentence order and the shuffled sentence order inherently.
Given that this tendency is not consistent with RoBERTa-base, the next sentence prediction task for pretraining BERT-base may cause this difference.

\subsection{Out-of-Distribution Generalization}
\label{sec:ood}

If models can recognize the intended features described in Table \ref{tb:perturbation} after entropy maximization, it is possible that the way models process language becomes closer to the way humans do, and thereby generalize to OOD \cite{talmor-berant-2019-multiqa} and adversarial \cite{jia-liang-2017-adversarial} test sets better than before.

The F1 scores on datasets from other domains and adversarial test sets are shown in Tables \ref{tb:ood-detail} and \ref{tb:adversarial-robustness}, respectively.
Based on these results, entropy maximization for any perturbations did not improve the generalization to other domains nor the adversarial robustness, but rather sometimes degraded them.

As discussed in \citet{hase2021search}, intentionally perturbed inputs can be out-of-distribution for models, and do not naturally appear in a dataset.
Therefore, regularizing the predictions on largely perturbed inputs may not have a positive effect on the generalization to natural examples.
Making QA models recognize natural changes in inputs using carefully designed perturbations (e.g., masked language modeling \cite{bert}) is future work.

\section{Related Work}
\subsection{Insensitivity to Large Perturbations}
Recently, there has been a surge of interest in the insensitivity of NLU models to large perturbations \cite{sugawara-etal-2018-makes,sugawara-etal-2020-assessing,hessel-schofield-2021-effective}.
These studies showed that NLU models can correctly predict the output even when the inputs is largely perturbed with some transformations such as word deletion and word order shuffling at test time.
To penalize the insensitivity of NLU models, entropy maximization has been used \cite{feng-etal-2018-pathologies,gupta_bert_2021,sinha-etal-2021-unnatural}.
However, the effect of entropy maximization for large perturbations has been studied in isolation.
Given the hypothesis that different perturbation removes different features from input as shown in Table \ref{tb:perturbation}, only regularizing entropy for single type of perturbation may not be enough to make models' predictions more human-like.

Making dialog models recognize dialog history with perturbations has been shown to improve the performance of dialog systems \cite{zhou-etal-2021-learning}.
This work is most relevant to our motivation.

\subsection{Sensitivity to Small Perturbations}
In contrast, deep learning models can misclassify slightly perturbed inputs \cite{szegedy2013intriguing,jia-liang-2017-adversarial,mudrakarta-etal-2018-model}, which are called adversarial examples.
To mitigate this issue, adversarial training and its variants have been proven to be effective \cite{goodfellow2014explaining,Zhu2020FreeLB,jiang-etal-2020-smart,liu2020adversarial}
in maintaining the model prediction when the input is slightly perturbed, which is the exact opposite of entropy maximization used in our work.
Similar to our work, the transfer of adversarial training between different perturbation types was studied  in computer vision \cite{kang2019transfer,pmlr-v119-maini20a}.

\section{Conclusion}
We first showed that entropy maximization often fails to transfer to unseen perturbation types.
Maximizing the entropy terms for various types of perturbations is effective in mitigating this problem.
The failure of entropy maximization to improve out-of-distribution generalization may be caused by the unnaturalness of the perturbed inputs.
Modifying the perturbation functions to effectively improve out-of-distribution generalization is future work.

\section*{Acknowledgements}
We would like to thank the anonymous reviewers for their valuable comments.
This work was supported by JST SPRING Grant Number JPMJSP2108 and JSPS KAKENHI Grant Numbers 21H03502, 22J13751 and 22K17954.
This work was also supported by NEDO SIP-2 ``Big-data and AI-enabled Cyberspace Technologies''.

\bibliography{anthology,custom}

\begin{thebibliography}{44}
\providecommand{\natexlab}[1]{#1}

\bibitem[{Belinkov(2022)}]{belinkov-2022-probing}
Belinkov, Y. 2022.
\newblock Probing Classifiers: Promises, Shortcomings, and Advances.
\newblock \emph{Computational Linguistics}, 48(1): 207--219.

\bibitem[{Brown et~al.(2020)Brown, Mann, Ryder, Subbiah, Kaplan, Dhariwal,
  Neelakantan, Shyam, Sastry, Askell, Agarwal, Herbert-Voss, Krueger, Henighan,
  Child, Ramesh, Ziegler, Wu, Winter, Hesse, Chen, Sigler, Litwin, Gray, Chess,
  Clark, Berner, McCandlish, Radford, Sutskever, and
  Amodei}]{brown-etal-2020-language}
Brown, T.; Mann, B.; Ryder, N.; Subbiah, M.; Kaplan, J.~D.; Dhariwal, P.;
  Neelakantan, A.; Shyam, P.; Sastry, G.; Askell, A.; Agarwal, S.;
  Herbert-Voss, A.; Krueger, G.; Henighan, T.; Child, R.; Ramesh, A.; Ziegler,
  D.; Wu, J.; Winter, C.; Hesse, C.; Chen, M.; Sigler, E.; Litwin, M.; Gray,
  S.; Chess, B.; Clark, J.; Berner, C.; McCandlish, S.; Radford, A.; Sutskever,
  I.; and Amodei, D. 2020.
\newblock Language Models are Few-Shot Learners.
\newblock In Larochelle, H.; Ranzato, M.; Hadsell, R.; Balcan, M.~F.; and Lin,
  H., eds., \emph{Advances in Neural Information Processing Systems},
  volume~33, 1877--1901. Curran Associates, Inc.

\bibitem[{Devlin et~al.(2019)Devlin, Chang, Lee, and Toutanova}]{bert}
Devlin, J.; Chang, M.-W.; Lee, K.; and Toutanova, K. 2019.
\newblock {BERT}: Pre-training of Deep Bidirectional Transformers for Language
  Understanding.
\newblock In \emph{Proceedings of the 2019 Conference of the North {A}merican
  Chapter of the Association for Computational Linguistics: Human Language
  Technologies, Volume 1 (Long and Short Papers)}, 4171--4186. Minneapolis,
  Minnesota: Association for Computational Linguistics.

\bibitem[{DeYoung et~al.(2020)DeYoung, Jain, Rajani, Lehman, Xiong, Socher, and
  Wallace}]{deyoung-etal-2020-eraser}
DeYoung, J.; Jain, S.; Rajani, N.~F.; Lehman, E.; Xiong, C.; Socher, R.; and
  Wallace, B.~C. 2020.
\newblock {ERASER}: {A} Benchmark to Evaluate Rationalized {NLP} Models.
\newblock In \emph{Proceedings of the 58th Annual Meeting of the Association
  for Computational Linguistics}, 4443--4458. Online: Association for
  Computational Linguistics.

\bibitem[{Dolan and Brockett(2005)}]{dolan-brockett-2005-automatically}
Dolan, W.~B.; and Brockett, C. 2005.
\newblock Automatically Constructing a Corpus of Sentential Paraphrases.
\newblock In \emph{Proceedings of the Third International Workshop on
  Paraphrasing ({IWP}2005)}.

\bibitem[{Dunn et~al.(2017)Dunn, Sagun, Higgins, Guney, Cirik, and
  Cho}]{dunn2017searchqa}
Dunn, M.; Sagun, L.; Higgins, M.; Guney, V.~U.; Cirik, V.; and Cho, K. 2017.
\newblock Searchqa: A new q\&a dataset augmented with context from a search
  engine.
\newblock \emph{arXiv preprint arXiv:1704.05179}.

\bibitem[{Feng et~al.(2018)Feng, Wallace, Grissom~II, Iyyer, Rodriguez, and
  Boyd-Graber}]{feng-etal-2018-pathologies}
Feng, S.; Wallace, E.; Grissom~II, A.; Iyyer, M.; Rodriguez, P.; and
  Boyd-Graber, J. 2018.
\newblock Pathologies of Neural Models Make Interpretations Difficult.
\newblock In \emph{Proceedings of the 2018 Conference on Empirical Methods in
  Natural Language Processing}, 3719--3728. Brussels, Belgium: Association for
  Computational Linguistics.

\bibitem[{Fisch et~al.(2019)Fisch, Talmor, Jia, Seo, Choi, and
  Chen}]{fisch-etal-2019-mrqa}
Fisch, A.; Talmor, A.; Jia, R.; Seo, M.; Choi, E.; and Chen, D. 2019.
\newblock {MRQA} 2019 Shared Task: Evaluating Generalization in Reading
  Comprehension.
\newblock In \emph{Proceedings of the 2nd Workshop on Machine Reading for
  Question Answering}, 1--13. Hong Kong, China: Association for Computational
  Linguistics.

\bibitem[{Goodfellow, Shlens, and Szegedy(2014)}]{goodfellow2014explaining}
Goodfellow, I.~J.; Shlens, J.; and Szegedy, C. 2014.
\newblock Explaining and harnessing adversarial examples.
\newblock \emph{arXiv preprint arXiv:1412.6572}.

\bibitem[{Gupta, Kvernadze, and Srikumar(2021)}]{gupta_bert_2021}
Gupta, A.; Kvernadze, G.; and Srikumar, V. 2021.
\newblock {BERT} \&amp; {Family} {Eat} {Word} {Salad}: {Experiments} with
  {Text} {Understanding}.
\newblock \emph{Proceedings of the AAAI Conference on Artificial Intelligence},
  35(14): 12946--12954.

\bibitem[{Hase, Xie, and Bansal(2021)}]{hase2021search}
Hase, P.; Xie, H.; and Bansal, M. 2021.
\newblock Search Methods for Sufficient, Socially-Aligned Feature Importance
  Explanations with In-Distribution Counterfactuals.
\newblock \emph{arXiv preprint arXiv:2106.00786}.

\bibitem[{Hessel and Schofield(2021)}]{hessel-schofield-2021-effective}
Hessel, J.; and Schofield, A. 2021.
\newblock How effective is {BERT} without word ordering? Implications for
  language understanding and data privacy.
\newblock In \emph{Proceedings of the 59th Annual Meeting of the Association
  for Computational Linguistics and the 11th International Joint Conference on
  Natural Language Processing (Volume 2: Short Papers)}, 204--211. Online:
  Association for Computational Linguistics.

\bibitem[{Hewitt and Liang(2019)}]{hewitt-liang-2019-designing}
Hewitt, J.; and Liang, P. 2019.
\newblock Designing and Interpreting Probes with Control Tasks.
\newblock In \emph{Proceedings of the 2019 Conference on Empirical Methods in
  Natural Language Processing and the 9th International Joint Conference on
  Natural Language Processing (EMNLP-IJCNLP)}, 2733--2743. Hong Kong, China:
  Association for Computational Linguistics.

\bibitem[{Hewitt and Manning(2019)}]{hewitt-manning-2019-structural}
Hewitt, J.; and Manning, C.~D. 2019.
\newblock {A} Structural Probe for Finding Syntax in Word Representations.
\newblock In \emph{Proceedings of the 2019 Conference of the North {A}merican
  Chapter of the Association for Computational Linguistics: Human Language
  Technologies, Volume 1 (Long and Short Papers)}, 4129--4138. Minneapolis,
  Minnesota: Association for Computational Linguistics.

\bibitem[{Jia and Liang(2017)}]{jia-liang-2017-adversarial}
Jia, R.; and Liang, P. 2017.
\newblock Adversarial Examples for Evaluating Reading Comprehension Systems.
\newblock In \emph{Proceedings of the 2017 Conference on Empirical Methods in
  Natural Language Processing}, 2021--2031. Copenhagen, Denmark: Association
  for Computational Linguistics.

\bibitem[{Jiang et~al.(2020)Jiang, He, Chen, Liu, Gao, and
  Zhao}]{jiang-etal-2020-smart}
Jiang, H.; He, P.; Chen, W.; Liu, X.; Gao, J.; and Zhao, T. 2020.
\newblock {SMART}: Robust and Efficient Fine-Tuning for Pre-trained Natural
  Language Models through Principled Regularized Optimization.
\newblock In \emph{Proceedings of the 58th Annual Meeting of the Association
  for Computational Linguistics}, 2177--2190. Online: Association for
  Computational Linguistics.

\bibitem[{Joshi et~al.(2017)Joshi, Choi, Weld, and
  Zettlemoyer}]{joshi-etal-2017-triviaqa}
Joshi, M.; Choi, E.; Weld, D.; and Zettlemoyer, L. 2017.
\newblock {T}rivia{QA}: A Large Scale Distantly Supervised Challenge Dataset
  for Reading Comprehension.
\newblock In \emph{Proceedings of the 55th Annual Meeting of the Association
  for Computational Linguistics (Volume 1: Long Papers)}, 1601--1611.
  Vancouver, Canada: Association for Computational Linguistics.

\bibitem[{Kang et~al.(2019)Kang, Sun, Brown, Hendrycks, and
  Steinhardt}]{kang2019transfer}
Kang, D.; Sun, Y.; Brown, T.; Hendrycks, D.; and Steinhardt, J. 2019.
\newblock Transfer of adversarial robustness between perturbation types.
\newblock \emph{arXiv preprint arXiv:1905.01034}.

\bibitem[{Kingma and Ba(2014)}]{kingma2014adam}
Kingma, D.~P.; and Ba, J. 2014.
\newblock Adam: A method for stochastic optimization.
\newblock \emph{arXiv preprint arXiv:1412.6980}.

\bibitem[{Kwiatkowski et~al.(2019)Kwiatkowski, Palomaki, Redfield, Collins,
  Parikh, Alberti, Epstein, Polosukhin, Devlin, Lee, Toutanova, Jones, Kelcey,
  Chang, Dai, Uszkoreit, Le, and Petrov}]{kwiatkowski-etal-2019-natural}
Kwiatkowski, T.; Palomaki, J.; Redfield, O.; Collins, M.; Parikh, A.; Alberti,
  C.; Epstein, D.; Polosukhin, I.; Devlin, J.; Lee, K.; Toutanova, K.; Jones,
  L.; Kelcey, M.; Chang, M.-W.; Dai, A.~M.; Uszkoreit, J.; Le, Q.; and Petrov,
  S. 2019.
\newblock Natural Questions: A Benchmark for Question Answering Research.
\newblock \emph{Transactions of the Association for Computational Linguistics},
  7: 452--466.

\bibitem[{Lewis et~al.(2020)Lewis, Liu, Goyal, Ghazvininejad, Mohamed, Levy,
  Stoyanov, and Zettlemoyer}]{lewis-etal-2020-bart}
Lewis, M.; Liu, Y.; Goyal, N.; Ghazvininejad, M.; Mohamed, A.; Levy, O.;
  Stoyanov, V.; and Zettlemoyer, L. 2020.
\newblock {BART}: Denoising Sequence-to-Sequence Pre-training for Natural
  Language Generation, Translation, and Comprehension.
\newblock In \emph{Proceedings of the 58th Annual Meeting of the Association
  for Computational Linguistics}, 7871--7880. Online: Association for
  Computational Linguistics.

\bibitem[{Li, Monroe, and Jurafsky(2016)}]{li2016understanding}
Li, J.; Monroe, W.; and Jurafsky, D. 2016.
\newblock Understanding neural networks through representation erasure.
\newblock \emph{arXiv preprint arXiv:1612.08220}.

\bibitem[{Liu et~al.(2020)Liu, Cheng, He, Chen, Wang, Poon, and
  Gao}]{liu2020adversarial}
Liu, X.; Cheng, H.; He, P.; Chen, W.; Wang, Y.; Poon, H.; and Gao, J. 2020.
\newblock Adversarial training for large neural language models.
\newblock \emph{arXiv preprint arXiv:2004.08994}.

\bibitem[{Liu et~al.(2019)Liu, Ott, Goyal, Du, Joshi, Chen, Levy, Lewis,
  Zettlemoyer, and Stoyanov}]{liu2019roberta}
Liu, Y.; Ott, M.; Goyal, N.; Du, J.; Joshi, M.; Chen, D.; Levy, O.; Lewis, M.;
  Zettlemoyer, L.; and Stoyanov, V. 2019.
\newblock Roberta: A robustly optimized bert pretraining approach.
\newblock \emph{arXiv preprint arXiv:1907.11692}.

\bibitem[{Maini, Wong, and Kolter(2020)}]{pmlr-v119-maini20a}
Maini, P.; Wong, E.; and Kolter, Z. 2020.
\newblock Adversarial Robustness Against the Union of Multiple Perturbation
  Models.
\newblock In III, H.~D.; and Singh, A., eds., \emph{Proceedings of the 37th
  International Conference on Machine Learning}, volume 119 of
  \emph{Proceedings of Machine Learning Research}, 6640--6650. PMLR.

\bibitem[{Mudrakarta et~al.(2018)Mudrakarta, Taly, Sundararajan, and
  Dhamdhere}]{mudrakarta-etal-2018-model}
Mudrakarta, P.~K.; Taly, A.; Sundararajan, M.; and Dhamdhere, K. 2018.
\newblock Did the Model Understand the Question?
\newblock In \emph{Proceedings of the 56th Annual Meeting of the Association
  for Computational Linguistics (Volume 1: Long Papers)}, 1896--1906.
  Melbourne, Australia: Association for Computational Linguistics.

\bibitem[{Pearl(2000)}]{pearl2000models}
Pearl, J. 2000.
\newblock Models, reasoning and inference.
\newblock \emph{Cambridge, UK: CambridgeUniversityPress}, 19.

\bibitem[{Radford et~al.(2019)Radford, Wu, Child, Luan, Amodei, Sutskever
  et~al.}]{radford2019language}
Radford, A.; Wu, J.; Child, R.; Luan, D.; Amodei, D.; Sutskever, I.; et~al.
  2019.
\newblock Language models are unsupervised multitask learners.
\newblock \emph{OpenAI blog}.

\bibitem[{Rajpurkar et~al.(2016)Rajpurkar, Zhang, Lopyrev, and
  Liang}]{rajpurkar-etal-2016-squad}
Rajpurkar, P.; Zhang, J.; Lopyrev, K.; and Liang, P. 2016.
\newblock {SQ}u{AD}: 100,000+ Questions for Machine Comprehension of Text.
\newblock In \emph{Proceedings of the 2016 Conference on Empirical Methods in
  Natural Language Processing}, 2383--2392. Austin, Texas: Association for
  Computational Linguistics.

\bibitem[{Sen and Saffari(2020)}]{sen-saffari-2020-models}
Sen, P.; and Saffari, A. 2020.
\newblock What do Models Learn from Question Answering Datasets?
\newblock In \emph{Proceedings of the 2020 Conference on Empirical Methods in
  Natural Language Processing (EMNLP)}, 2429--2438. Online: Association for
  Computational Linguistics.

\bibitem[{Seo et~al.(2017)Seo, Kembhavi, Farhadi, and
  Hajishirzi}]{seo2017bidirectional}
Seo, M.; Kembhavi, A.; Farhadi, A.; and Hajishirzi, H. 2017.
\newblock Bidirectional Attention Flow for Machine Comprehension.
\newblock In \emph{International Conference on Learning Representations}.

\bibitem[{Shannon(1948)}]{shannon1948mathematical}
Shannon, C.~E. 1948.
\newblock A mathematical theory of communication.
\newblock \emph{The Bell system technical journal}, 27(3): 379--423.

\bibitem[{Sinha et~al.(2021)Sinha, Parthasarathi, Pineau, and
  Williams}]{sinha-etal-2021-unnatural}
Sinha, K.; Parthasarathi, P.; Pineau, J.; and Williams, A. 2021.
\newblock {UnNatural} {L}anguage {I}nference.
\newblock In \emph{Proceedings of the 59th Annual Meeting of the Association
  for Computational Linguistics and the 11th International Joint Conference on
  Natural Language Processing (Volume 1: Long Papers)}, 7329--7346. Online:
  Association for Computational Linguistics.

\bibitem[{Sugawara et~al.(2018)Sugawara, Inui, Sekine, and
  Aizawa}]{sugawara-etal-2018-makes}
Sugawara, S.; Inui, K.; Sekine, S.; and Aizawa, A. 2018.
\newblock What Makes Reading Comprehension Questions Easier?
\newblock In \emph{Proceedings of the 2018 Conference on Empirical Methods in
  Natural Language Processing}, 4208--4219. Brussels, Belgium: Association for
  Computational Linguistics.

\bibitem[{Sugawara et~al.(2020)Sugawara, Stenetorp, Inui, and
  Aizawa}]{sugawara-etal-2020-assessing}
Sugawara, S.; Stenetorp, P.; Inui, K.; and Aizawa, A. 2020.
\newblock Assessing the {Benchmarking} {Capacity} of {Machine} {Reading}
  {Comprehension} {Datasets}.
\newblock \emph{Proceedings of the AAAI Conference on Artificial Intelligence},
  34(05): 8918--8927.

\bibitem[{Szegedy et~al.(2013)Szegedy, Zaremba, Sutskever, Bruna, Erhan,
  Goodfellow, and Fergus}]{szegedy2013intriguing}
Szegedy, C.; Zaremba, W.; Sutskever, I.; Bruna, J.; Erhan, D.; Goodfellow, I.;
  and Fergus, R. 2013.
\newblock Intriguing properties of neural networks.
\newblock \emph{arXiv preprint arXiv:1312.6199}.

\bibitem[{Talmor and Berant(2019)}]{talmor-berant-2019-multiqa}
Talmor, A.; and Berant, J. 2019.
\newblock {M}ulti{QA}: An Empirical Investigation of Generalization and
  Transfer in Reading Comprehension.
\newblock In \emph{Proceedings of the 57th Annual Meeting of the Association
  for Computational Linguistics}, 4911--4921. Florence, Italy: Association for
  Computational Linguistics.

\bibitem[{Tenney et~al.(2019)Tenney, Xia, Chen, Wang, Poliak, McCoy, Kim,
  Durme, Bowman, Das, and Pavlick}]{tenney2018what}
Tenney, I.; Xia, P.; Chen, B.; Wang, A.; Poliak, A.; McCoy, R.~T.; Kim, N.;
  Durme, B.~V.; Bowman, S.; Das, D.; and Pavlick, E. 2019.
\newblock What do you learn from context? Probing for sentence structure in
  contextualized word representations.
\newblock In \emph{International Conference on Learning Representations}.

\bibitem[{Trischler et~al.(2017)Trischler, Wang, Yuan, Harris, Sordoni,
  Bachman, and Suleman}]{trischler-etal-2017-newsqa}
Trischler, A.; Wang, T.; Yuan, X.; Harris, J.; Sordoni, A.; Bachman, P.; and
  Suleman, K. 2017.
\newblock {N}ews{QA}: A Machine Comprehension Dataset.
\newblock In \emph{Proceedings of the 2nd Workshop on Representation Learning
  for {NLP}}, 191--200. Vancouver, Canada: Association for Computational
  Linguistics.

\bibitem[{Williams, Nangia, and Bowman(2018)}]{williams-etal-2018-broad}
Williams, A.; Nangia, N.; and Bowman, S. 2018.
\newblock A Broad-Coverage Challenge Corpus for Sentence Understanding through
  Inference.
\newblock In \emph{Proceedings of the 2018 Conference of the North {A}merican
  Chapter of the Association for Computational Linguistics: Human Language
  Technologies, Volume 1 (Long Papers)}, 1112--1122. New Orleans, Louisiana:
  Association for Computational Linguistics.

\bibitem[{Yang et~al.(2018)Yang, Qi, Zhang, Bengio, Cohen, Salakhutdinov, and
  Manning}]{yang-etal-2018-hotpotqa}
Yang, Z.; Qi, P.; Zhang, S.; Bengio, Y.; Cohen, W.; Salakhutdinov, R.; and
  Manning, C.~D. 2018.
\newblock {H}otpot{QA}: A Dataset for Diverse, Explainable Multi-hop Question
  Answering.
\newblock In \emph{Proceedings of the 2018 Conference on Empirical Methods in
  Natural Language Processing}, 2369--2380. Brussels, Belgium: Association for
  Computational Linguistics.

\bibitem[{Yogatama et~al.(2019)Yogatama, d'Autume, Connor, Kocisky,
  Chrzanowski, Kong, Lazaridou, Ling, Yu, Dyer et~al.}]{yogatama2019learning}
Yogatama, D.; d'Autume, C. d.~M.; Connor, J.; Kocisky, T.; Chrzanowski, M.;
  Kong, L.; Lazaridou, A.; Ling, W.; Yu, L.; Dyer, C.; et~al. 2019.
\newblock Learning and evaluating general linguistic intelligence.
\newblock \emph{arXiv preprint arXiv:1901.11373}.

\bibitem[{Zhou, Li, and Li(2021)}]{zhou-etal-2021-learning}
Zhou, W.; Li, Q.; and Li, C. 2021.
\newblock Learning from Perturbations: Diverse and Informative Dialogue
  Generation with Inverse Adversarial Training.
\newblock In \emph{Proceedings of the 59th Annual Meeting of the Association
  for Computational Linguistics and the 11th International Joint Conference on
  Natural Language Processing (Volume 1: Long Papers)}, 694--703. Online:
  Association for Computational Linguistics.

\bibitem[{Zhu et~al.(2020)Zhu, Cheng, Gan, Sun, Goldstein, and
  Liu}]{Zhu2020FreeLB}
Zhu, C.; Cheng, Y.; Gan, Z.; Sun, S.; Goldstein, T.; and Liu, J. 2020.
\newblock FreeLB: Enhanced Adversarial Training for Natural Language
  Understanding.
\newblock In \emph{International Conference on Learning Representations}.

\end{thebibliography}
\end{document}